\documentclass[a4paper]{article}

\usepackage{INTERSPEECH2022}
\usepackage{amsmath,graphicx}
\usepackage{multirow, hhline, stmaryrd, url, cases, bbm}

\title{Internal Language Model Adaptation with Text-Only Data for End-to-End Speech Recognition}
\name{Zhong Meng$^1$, Yashesh Gaur$^1$, Naoyuki Kanda$^1$, Jinyu Li$^1$, Xie Chen$^1$, Yu Wu$^2$, Yifan Gong$^1$}
\address{$^1$Microsoft Corporation, USA \hspace{3mm} $^2$Microsoft Research Asia, China}
\email{\{zhme,yagaur,nakanda,jinyli,xieche,yuwu1,ygong\}@microsoft.com}

\begin{document}

\maketitle
\begin{abstract}
Text-only adaptation of an end-to-end (E2E) model remains a challenging task for automatic speech recognition (ASR). Language model (LM) fusion-based approaches require an additional external LM during inference, significantly increasing the computation cost. 
To overcome this, we propose an internal LM adaptation (ILMA) of the E2E model using \emph{text-only} data. Trained with audio-transcript pairs, an E2E model implicitly learns an internal LM that characterizes the token sequence probability which is approximated by the E2E model output after zeroing out the encoder contribution. During ILMA, we fine-tune the internal LM, i.e., the E2E components excluding the encoder, to minimize a cross-entropy loss. 
To make ILMA effective, it is essential to train the E2E model with an internal LM loss besides the standard E2E loss. Furthermore, we propose to regularize ILMA by minimizing the Kullback-Leibler divergence between the output distributions of the adapted and unadapted internal LMs. ILMA is the most effective when we update only the last linear layer of the joint network. ILMA enables a fast text-only adaptation of the E2E model without increasing the run-time computational cost. 
Experimented with 30K-hour trained transformer transducer models, ILMA achieves up to 34.9\% relative word error rate reduction from the unadapted baseline.
\end{abstract}

\section{Introduction}
\label{sec:intro}
End-to-end (E2E) models have achieved the state-of-art performance for automatic speech recognition (ASR)  \cite{chiu2018state, sainath2020streaming, li2020developing, li2021recent}. With the goal of directly mapping speech features into word sequences using a single neural network (NN), the most popular E2E models include connectionist temporal classification \cite{graves2006connectionist, hannun2014deep, Li18CTCnoOOV}, recurrent neural network transducer (RNN-T) \cite{graves2012sequence, sainath2020streaming,  li2020developing}, and attention-based encoder-decoder (AED) models \cite{chorowski2015attention, karita2019comparative, Li2020Comparison}. However, the performance of E2E models degrades when a mismatch exists between the source-domain training data and the target-domain test data. 
Many ideas have been proposed to adapt ASR models, such as regularization methods \cite{kld_yu, meng2019asa,l2_liao, meng2020lvector, lhuc}, teacher-student learning \cite{li2014learning, meng2018adversarial, manohar2018teacher, meng2019conditional}, 
transformation methods \cite{lhn, tan2015cluster}, 
and adversarial learning \cite{meng2018speaker, grl_shinohara, grl_serdyuk, dsn_meng}. However, all these methods require audio data for adaptation when applied to E2E models \cite{ochiai2018speaker, meng2019speaker, meng2019domain}.
To overcome this, the most promising approach is to adapt the E2E model using \emph{text-only} data because it is easy to collect orders of magnitude more text-only data than the audio-transcript pairs in the target domain.

The most common solution is to train an LM using text-only adaptation data and integrate it into the E2E model during inference. The simplest LM fusion method is Shallow Fusion \cite{hannun2014deep, gulcehre2015on} which combines the E2E model score and the external LM score in the log-linear domain at each step of the beam search. To improve Shallow Fusion, a Density Ratio method \cite{mcdermott2019density, kanda2016maximum} and an internal LM estimation-based Fusion \cite{variani2020hybrid, meng2021ilme} were proposed in which a source-domain LM score and an internal LM score are subtracted from the Shallow Fusion score, respectively, during inference. 
Variants of internal LM estimation were proposed in \cite{zeyer2021librispeech, zeineldeen2021investigating}.
Minimum word error rate training with LM fusion \cite{meng2021minimum, kanda2017minimum, weng2019minimum, peyser2020improving} was conducted to obviate the need for LM weights tuning. 
Furthermore, structural LM fusion methods such as Deep Fusion \cite{gulcehre2015on}, Cold Fusion \cite{sriram2017cold} and Simple Fusion \cite{stahlberg2018simple} 
jointly train an E2E model with an external LM to learn a sophisticated combination between the two models.
However, all these fusion methods involve an additional external LM during inference which significantly increases the run-time computational cost.

Other solutions include synthesizing speech from the text-only adaptation data using a text-to-speech (TTS) model, and then fine-tuning the E2E model with the synthesized audio-transcript pairs \cite{huang2020rapid, peyser2020improving, baskar2019self}. During inference, only the adapted E2E model is needed.
However, these approaches require additional transcribed speech to train a TTS model. Training such a reliable multi-speaker TTS model and generating synthesized speech from it are both computationally expensive. Moreover, the adaptation cost with audio-transcript pairs is typically much higher than that with text-only data. 
Therefore, it is hard to apply these methods to the fast on-device adaptation 
where the target-domain text is frequently updated. 
Recently, \cite{pylkkonen2021fast} shows that an RNN-T can be adapted with text-only data by training an auxiliary LM output layer and using it as a regularizer to fine-tune the prediction network. However, learning the additional LM output layer complicates the adaptation process, leading to increased adaptation time and computational cost. \cite{chen2022factorized} factorizes the blank and vocabulary predictions in a transducer and  effectively adapts the vocabulary predictor with text as adapting an LM.  However, the additional blank predictor increases the model size and computational cost. 

To address these limitations, we propose 
an internal LM adaptation (ILMA)
of an E2E model using the \emph{text-only} data to improve the ASR performance on the target-domain test data. 
Trained with audio-transcript pairs, an E2E model implicitly learns an internal LM that characterizes token sequence distribution. The internal LM probability is estimated by the E2E model output after eliminating the encoder contribution from the network \cite{variani2020hybrid, meng2021ilme}. Therefore, we define the E2E model components excluding the encoder as an \emph{internal LM}. Specifically, the internal LM refers to the prediction network followed by the joint network for a transducer model, or to the decoder for an AED model. 
During ILMA, we fine-tune the internal LM parameters to maximize the log probability of the adaptation sentences. 
Before ILMA, it is essential to perform internal LM training (ILMT) \cite{meng2021ilmt} to minimize a cross-entropy internal LM loss in addition to the standard E2E loss. ILMT learns a \emph{dual-mode} internal LM that acts as a standalone LM without losing its capability of generating the correct E2E scores together with the other components of the E2E model. Kullback-Leibler divergence (KLD) \cite{kullback1951information} regularization is performed during ILMA to prevent the adapted E2E model from overfitting the text-only data.  We show that, for a transducer model, ILMA is the most effective when we update only the last linear layer of the joint network. Compared to LM fusion and TTS-based adaptation, ILMA enables a fast text-only adaptation without increasing run-time computational cost during inference. Evaluated with 30 thousand (K)-hour trained transformer transducers (T-Ts) \cite{zhang2020transformer}, ILMA achieves up to 34.9\% relative word error rate (WER) reduction from the unadapted T-T on Microsoft production test sets.


\section{Related Work}
\subsection{E2E Models for ASR}
Given a sequence of speech features $\mathbf{X}=\{\mathbf{x}_1,
\ldots, \mathbf{x}_T\}$ where $\mathbf{x}_t \in \mathbbm{R}^d$, an E2E model is trained to minimize the summation of the negative log posteriors of the reference token sequences $\mathbf{Y}=\{y_1, \ldots, y_U\}$ over the training corpus $\mathcal{D}_\text{T}$ where $y_u \in \mathcal{V}$ and $\mathcal{V}$ is the set of non-blank output tokens, e.g., word pieces,
\begin{align}
   \mathcal{L}_\text{E2E}(\theta_\text{E2E}) = - \sum_{(\mathbf{X}, \mathbf{Y}) \in \mathcal{D}_\text{T}} \log P(\mathbf{Y} | \mathbf{X};\theta_\text{E2E}), \label{eqn:e2e_loss}
\end{align}
where $\theta_\text{E2E}$ is the parameters of an E2E model.

A transducer model \cite{graves2012sequence} comprises an encoder, a prediction network and a joint network. The encoder maps the input speech features $\mathbf{X}$ to a sequence of hidden states $\mathbf{H}^{\text{enc}} = \{\mathbf{h}^\text{enc}_1, \ldots, \mathbf{h}^\text{enc}_T\}$. 
The predictor takes in the embedding vectors of the previous \emph{non-blank} token $\mathbf{Y}_{0:u-1}$ to generate the hidden state $\mathbf{h}^\text{pred}_u$. 
The joint network is a feed-forward network that combines the outputs of the encoder and predictor to predict the conditional distribution over the next possible token $k\in \mathcal{V} \cup \texttt{<b>}$, i.e., 
\begin{align}
    & \mathbf{z}_{t, u} = W_j \phi(W_e\mathbf{h}^{\text{enc}}_{t} + W_p\mathbf{h}^\text{pred}_{u} + \mathbf{b}_e + \mathbf{b}_p) + \mathbf{b}_j, \label{eqn:rnnt_logit} \\
    \hspace{-4pt} 
    & \left[P(k|t, u; \theta_\text{E2E})\right]_{k\in \mathcal{V} \cup \texttt{<b>}}= \text{softmax}(\mathbf{z}_{t, u}), \label{eqn:rnnt_softmax}
\end{align}
where $\texttt{<b>}$ denotes a blank symbol, $\phi$ is a non-linear function, e.g., tanh or ReLU. $W_j$, $W_e$, $W_p$ are weight matrices, and $\mathbf{b}_e$, $\mathbf{b}_p$, $\mathbf{b}_j$ are biases. $\mathbf{z}_{t, u}$ is a $|\mathcal{V}|+1$ dimensional logit vector. 

\subsection{Internal LM of E2E Model}
\label{sec:ilme}

The E2E model implicitly learns an internal LM via the E2E training with paired audio and transcripts. The internal LM characterizes the distribution of token sequences given $\mathbf{\theta}_\text{E2E}$. The internal LM probability of a token sequence $\mathbf{Y}$ is
\begin{align}
    & P(\mathbf{Y};\theta_\text{E2E}) = \prod^{U}_{u=1}P(y_u|\mathbf{Y}_{0:u-1};\theta_\text{E2E}), \label{eqn:e2e_ilm}
\end{align}
where $P(y_u|\mathbf{Y}_{0:u-1};\theta_\text{E2E})$ is estimated as the E2E model output at the step $u$ after zeroing out the encoder hidden states $\mathbf{H}^\text{enc}$ \cite{variani2020hybrid, meng2021ilme}. 
For a transducer, the internal LM probability is the model output after feeding the previous tokens $\mathbf{Y}_{0:u-1}$ through the internal LM, i.e., the prediction network followed by the joint network
\begin{align}
    & \mathbf{z}^\text{ILM, NB}_u = \mathbf{W}^\text{NB}_j\phi(\mathbf{W}_p\mathbf{h}^\text{pred}_u + \mathbf{b}_p) + \mathbf{b}^\text{NB}_j \label{eqn:logit_ilm}, \\
    & P(y_u|\mathbf{Y}_{0:u-1}; \theta_\text{pred}, \theta_\text{joint}) = \text{softmax}(\mathbf{z}^\text{ILM, NB}_{u}), \label{eqn:rnnt_cond_ilm}
\end{align}
where NB denotes non-blank, $\mathbf{W}_j^\text{NB}$ and $\mathbf{b}^\text{NB}_j$ are created from $\mathbf{W}_j$ and $\mathbf{b}_j$, respectively, by removing the rows corresponding to the blank, and $\mathbf{z}^\text{ILM, NB}_u$ is a logit vector of dimension $|\mathcal{V}|$ without the blank logit.
$\theta_\text{pred}$ and $\theta_\text{joint}$ are the parameters of the prediction network and joint network, respectively.

\subsection{Internal LM Training (ILMT)}
\label{sec:ilmt}

The standard E2E training learns a weak internal LM with a high perplexity because it maximizes the posterior $P(\mathbf{Y}|\mathbf{X};\theta_\text{E2E})$ instead of the internal LM probability $P(\mathbf{Y};\theta_\text{E2E})$. To address this limitation, ILMT was proposed in \cite{meng2021ilmt} where the E2E model is trained to minimize an internal LM loss in additional to the standard E2E loss.

The internal LM loss is the summation of negative log internal LM probabilities over the training corpus $\mathcal{D}_\text{T}$ as follows
\begin{align}
    \hspace{-2pt}\mathcal{L}_{\text{ILM}}(\theta_\text{ILM}; \mathcal{D}_\text{T}) = -\sum_{\mathbf{Y} \in \mathcal{D}_\text{T}} \sum^{U}_{u=1}\log P(y_u|\mathbf{Y}_{0:u-1};\theta_\text{ILM}), \label{eqn:ilm_loss_train}
\end{align}
where $\theta_\text{ILM}$ denotes the internal LM parameters. $\theta_\text{ILM}$ equals to $\{\theta_\text{pred}, \theta_\text{join}\}$ for a transducer model, and denotes
the decoder parameters for an AED model.
The ILMT loss is a weighted sum of the E2E loss in Eq. \eqref{eqn:e2e_loss} and the internal LM loss in Eq. \eqref{eqn:ilm_loss_train}
\begin{align}
    & \hspace{-3pt} \mathcal{L}_{\text{ILMT}}(\theta_\text{E2E}; \mathcal{D}_\text{T}) = \mathcal{L}_{\text{E2E}}(\theta_\text{E2E}; \mathcal{D}_\text{T}) + 
    \alpha \mathcal{L}_{\text{ILM}}(\theta_\text{ILM}; \mathcal{D}_\text{T}), \label{eqn:ilmt}
\end{align}
where $\alpha$ is the weight of the internal LM loss. By minimizing the ILMT loss, we maximize the internal LM probability of the E2E training transcripts by updating only the internal LM while maximizing the conditional probability of the training transcripts given input speech by updating the entire E2E model.

ILMT learns a strong internal LM with significantly reduced perplexity without losing the ASR performance. It encourages the internal LM to behave also like a standalone NN-LM \cite{mikolov2010recurrent}, increasing the modularity of the E2E model.

\section{Internal LM Adaptation (ILMA)}
Our goal is to adapt the E2E model using \emph{text-only} data so that it achieves improved ASR performance in the target domain without any increase of the run-time computational cost. 
Since an internal LM of an E2E model acts exactly the same as an standard NN-LM, one possible way is to adapt the internal LM using the text-only data. 
Theoretically, any NN-LM adaptation methods are applicable to ILMA.
The simplest solution is to re-train the internal LM to minimize the cross-entropy internal LM loss as follows 
\begin{align}
    & \hspace{-2pt}\mathcal{L}_{\text{ILM}}(\theta_\text{ILM};\mathcal{D}_\text{A}) = - \hspace{-2pt} \sum_{\mathbf{Y} \in \mathcal{D}_\text{A}} \sum^{U}_{u=1}\log P(y_u|\mathbf{Y}_{0:u-1};\theta_\text{ILM}). \label{eqn:ilm_loss_adapt}
\end{align}
Different from ILMT where the internal LM loss in Eq. \eqref{eqn:ilm_loss_train} is computed with the transcripts of the training data $\mathcal{D}_\text{T}$, the internal LM loss of ILMA is computed with the sentences in the text-only adaptation set $\mathcal{D}_\text{A}$.


In the standard E2E model training, the internal LM is updated together with the encoder to minimize the internal LM loss. However, in Eq. \eqref{eqn:ilm_loss_adapt}, the internal LM is re-trained separately to become a standalone NN-LM by maximizing the log probability of the adaptation sentences. After ILMA of a standard E2E model, it is probable that the adapted internal LM deviates too much from the unadapted one such that it forgets how to work together with the encoder to generate reasonable E2E model probabilities during inference. 
To circumvent this, we perform ILMT of the E2E model to minimize an ILMT loss in Eq. \eqref{eqn:ilmt} before adapting the internal LM towards text-only data. Through ILMT, 
a \emph{dual-mode} internal LM is learned to behave like a standalone LM while maintaining its capability to collaborate with the encoder to compute the correct E2E model probabilities. During ILMA, only the ``LM mode'' of the internal LM is adapted towards the text-only data while its ``E2E component mode'' remains unchanged, leading to improved performance on the target-domain test data. 

Further, to prevent the adapted E2E model from overfitting the text-only adaptation data, we minimize the KLD between the output distribution of the adapted internal LM and that of unadapted one below in addition to the cross-entropy internal LM loss
\begin{align}
    & \mathcal{KL}\left[P(y_u|\mathbf{Y}_{0:u-1};\theta^*_\text{ILM}) || P(y_u|\mathbf{Y}_{0:u-1};\theta_\text{ILM})\right] = \nonumber \\
    & \sum_{v \in\mathcal{V}} P( v|\mathbf{Y}_{0:u-1};\theta^*_\text{ILM}) \log \left[\frac{P( v|\mathbf{Y}_{0:u-1};\theta^*_\text{ILM})} {P( v|\mathbf{Y}_{0:u-1};\theta_\text{ILM})} \right],
    \label{eqn:kl}
\end{align}
where $\theta^*_\text{ILM}$ is the internal LM parameters before ILMA.
With KLD regularization, the ILMA loss becomes 
\begin{align}
    & \mathcal{L}_{\text{ILMA}}(\theta_\text{ILM}; \mathcal{D}_\text{A}) = (1 - \rho) \mathcal{L}_{\text{ILM}}(\theta_\text{ILM}; \mathcal{D}_\text{A}) \nonumber \\
    & - \rho \hspace{-2pt} \sum_{\mathbf{Y} \in \mathcal{D}_\text{A}} \sum_{u = 1}^U \sum_{v \in\mathcal{V}} P( v|\mathbf{Y}_{0:u-1};\theta^*_\text{ILM}) \log P( v|\mathbf{Y}_{0:u-1};\theta_\text{ILM}) \nonumber \\
    & = - \hspace{-5pt}\sum_{\mathbf{Y} \in \mathcal{D}_\text{A}} \sum_{u = 1}^U \sum_{v \in\mathcal{V}} \left[(1- \rho) \delta(v = y_u)  + \rho P( v|\mathbf{Y}_{0:u-1};\theta^*_\text{ILM}) \right] \nonumber \\
    & \qquad\qquad\qquad\qquad\quad   \log P(v|\mathbf{Y}_{0:u-1};\theta_\text{ILM}), \label{eqn:ilma}
\end{align}
where $\rho \in [0, 1]$ is the regularization weight and $\delta(v = y_u)$ is the Kronecker function which equals 1 when $v= y_u$ and $0$ otherwise. 
Note that ILMA with KLD regularization in Eq. \eqref{eqn:ilma} reduces to the cross-entropy ILMA in Eq. \eqref{eqn:ilm_loss_adapt} when $\rho = 0$. 

During ILMA, different parts of the internal LM parameters can be updated to minimize the ILMA loss. 
We find that, for a transducer model, 
ILMA is the most effective when updating only $\{\mathbf{W}^\text{NB}_j, \mathbf{b}_j^\text{NB}\}$ of the joint network. 
One possible reason is: The adaptation text does not include the blank token \texttt{<b>}, so the best we can do is to \emph{only} adapt the transducer’s capability of predicting non-blank tokens \emph{without} sacrificing its capability of predicting \texttt{<b>}. In a transducer, the only parameters that do not contribute to $P(\texttt{<b>}|\mathbf{Y}_{0:u-1};\theta^\text{S}_\text{E2E})$ are $\{\mathbf{W}_j^\text{NB}, \mathbf{b}_j^\text{NB}\}$. Therefore, adapting $\{\mathbf{W}_j^\text{NB}, \mathbf{b}_j^\text{NB}\}$ performs the best while adapting any other parameters may hurt the \texttt{<b>} predictions. 

In summary, ILMA of an E2E model with text-only data includes the following steps:
\begin{enumerate}
    \item Perform ILMT of the E2E model with audio-transcript pairs in the training set $\mathcal{D}_T$ by minimizing the ILMT loss in Eq. \eqref{eqn:ilmt} to generate the model $M_1$.
    \item Perform ILMA of the E2E model $M_1$ with \emph{text-only} data $\mathcal{D}_A$ by minimizing the ILMA loss defined in Eq. \eqref{eqn:ilma} to generate the model $M_2$.
    \item Perform inference with the adapted E2E model $M_2$ on the target-domain test data.
\end{enumerate}

ILMA enables a fast text-only adaptation of the internal LM 
without losing its functionality to work seamlessly with the other components of an E2E model. 
Contrary to LM fusion methods, ILMA does \emph{not} increase any computational cost during inference because only the adapted E2E model is needed to run the beam search decoding.
Compared to Fast Text-only Adaptation (FTA) \cite{pylkkonen2021fast}, ILMA saves one step of training the additional output LM component, effectively reducing the training time and the computational cost. Most importantly, ILMA has the flexibility to adapt only the joint network (the most effective) while FTA does not.

\section{Experiments}
We perform text-only ILMA of the T-T models trained with a transducer loss or a ILMT loss and evaluate the adapted models on intra-domain and cross-domain test sets.
During inference, we perform beam search with a beam size of 5 \emph{without} using any external LM. 3999 word-piece units generated by byte-pair encoding \cite{sennrich2015neural} are used as the output token set $\mathcal{V}$ for all T-T models. 

\subsection{Data Preparation}
We train T-T models with 30K hours of anonymized and transcribed data as in \cite{meng2021ilme, meng2021ilmt} collected from Microsoft services,
including voice search, short message dictation, conversations, etc. 
In addition, we collect two \emph{text-only} adaptation sets as follows
\begin{itemize}
    \item \textbf{Command \& Control}: 50 million (M) words of simulated text generated by a command \& control grammar and 400K words of real anonymized text from Microsoft command \& control services.
    \item \textbf{Multi-Domain}: 2 billion (B) words of anonymized text corpus comprising conversational data, voice search and short message dictation from Microsoft services.
\end{itemize}
Correspondingly, we gather two target-domain test sets whose transcripts do not overlap with the adaptation text.
\begin{itemize}
    \item \textbf{Command \& Control}: 1K utterances collected from Microsoft command \& control services.
    \item \textbf{Voice Search}: 18K far-field voice search utterances collected from Microsoft services. 
\end{itemize}
Note that Command \& Control is a \emph{cross-domain} evaluation set because it is not covered by the 30K-hour training data while Voice Search is an \emph{intra-domain} one because 30K-hour training data includes voice search utterances.
We extract 80-dimensional (dim) log Mel
filter bank features from the speech signal for training and test sets every 10 ms over a 25 ms window.

\subsection{Baseline Systems}
\label{sec:tt}

T-T has a streaming encoder \cite{chen2020developing} with four 2D convolution layers sub-sampling the 80-dim log Mel filter bank features in time by a factor of 4.
On top of that is a transformer with 18 layers, each layer has an 8-head attention sub-layer with relative positional encoding \cite{vaswani2017attention} and a 2048-dim fully-connected sub-layer. The encoder has a look-ahead of 360 ms on average. The T-T prediction network is a 2-layer
transformer with each layer containing a 4-head attention sub-layer followed by a 1024-dim fully-connected sub-layer. 
The input to the prediction network are 512-dim word-piece embedding vectors with positional encoding. 
The attention dimension is fixed at 512. The outputs of encoder and prediction network are projected to 512-dim vectors.
Dropout with a probability of 0.1 is deployed. T-T has 67 M parameters. 


We first train a T-T model to minimize the transducer loss using the 30K-hour data as the baseline model. In Table \ref{table:compare_adaptation}, the baseline T-T achieves 12.81\% and 4.23\% WERs on Command \& Control and Voice Search test sets, respectively.
To achieve an effective ILMA, we also perform the ILMT of T-T to minimize the ILMT loss in Eq. \eqref{eqn:ilmt} with the 30K-hour data before ILMA. 
The ILMT-ed T-T achieves 12.55\% and 4.22\% WERs on Command \& Control and Voice Search test sets, performing slightly better than the baseline T-T. With ILMT, the token-level perplexity of the internal LM reduces from 131.2 (baseline) to 60.7 on the validation set of the 30K-hour training data. This shows that a dual-mode internal LM with significantly lower perplexity and slightly better ASR performance is learned via ILMT.

To compare with ILMA, we perform FTA \cite{pylkkonen2021fast} on the baseline T-T. The output LM component is a feed-forward layer followed by a softmax. 
In Table \ref{table:compare_adaptation}, FTA achieves 10.04\% and 4.15\% WERs on Command \& Control and Voice Search test sets, respectively.
Moreover, we train two long short-term memory (LSTM) \cite{sak2014long,meng2017deep}-LMs with Command \& Control and Multi-Domain adaptation text, respectively. Both LSTM-LMs have two hidden layers and 2048 hidden units at each layer. Each LSTM-LM consists of 58M parameters. We perform Shallow Fusion of the ILMT-ed T-T and the LSTM-LM during inference. In Table \ref{table:compare_adaptation}, Shallow Fusion achieves 8.71\% and 4.04\% WERs for cross-domain and intra-domain evaluations, respectively.

\begin{table}
\centering
\setlength{\tabcolsep}{4.0pt}
\begin{tabular}[c]{c|c|c|c|c}
	\hline
	\hline
	\multirow{2}{*}{\begin{tabular}{@{}c@{}} Method \end{tabular}}
	& \multirow{2}{*}{\begin{tabular}{@{}c@{}} External \\ LM \end{tabular}} 
	& \multirow{2}{*}{\begin{tabular}{@{}c@{}} Model \\ \hspace{0.6pt} Params \hspace{0.6pt}\end{tabular}} 
	& \multirow{2}{*}{\begin{tabular}{@{}c@{}} Comd. \\ \hspace{0.1pt} Control \hspace{0.1pt} \end{tabular}} 
	& \multirow{2}{*}{\begin{tabular}{@{}c@{}} Voice \\ \hspace{1.1pt}Search\hspace{1.1pt} \end{tabular}} \\
	& & & & \\
	\hline
    T-T & 
	\multirow{3}{*}{\begin{tabular}{@{}c@{}} No \end{tabular}}
    & 
	\multirow{3}{*}{\begin{tabular}{@{}c@{}} 67M \end{tabular}}  
    & 12.81 & 4.23 \\
	\hhline{-~~--}
    ILMT & & & 12.55 & 4.22 \\
	\hhline{-~~--}
    FTA & & & 10.04 & 4.15 \\
	\hline
	ILMT + SF & Yes & 125M & 8.71 & 4.04 \\
	\hline
	ILMT + ILMA & No & 67M & \textbf{8.34} & \textbf{3.95} \\
	\hline
	\hline
	\end{tabular}
	\caption{Compare WERs (\%) and model sizes of 3 adaptation methods: Fast Text-only Adaptation (FTA) \cite{pylkkonen2021fast}, Shallow Fusion (SF) of ILMT-ed T-T and NN-LM, proposed ILMA after ILMT.}
\label{table:compare_adaptation}
\vspace{-10 pt}
\end{table}

\subsection{Internal LM Adaptation}
\label{sec:exp_ilma}
\subsubsection{Cross-Domain Evaluation}
\label{sec:exp_ilma_tt}

We adapt the baseline T-T with text-only data in the Command \& Control adaptation set by minimizing the ILMA loss in Eq. \eqref{eqn:ilma}.  Different components of the internal LM are updated during ILMA. 
We also adjust the KLD regularization weight $\rho$ to prevent the T-T from overfitting the adaptation text. As in Table \ref{table:command_control}, when updating the entire internal LM, the prediction network alone or $\{\mathbf{W}_j^\text{NB}, \mathbf{b}_j^\text{NB}\}$ of the joint network (Eq. \eqref{eqn:logit_ilm}), we obtain the largest relative WER reductions 6.8\%, 6.2\% and 19.1\%, respectively, from the baseline T-T at a common weight $\rho=0.2$. 
Adapting $\{\mathbf{W}_j^\text{NB}, \mathbf{b}_j^\text{NB}\}$ alone performs significantly better than 
adapting the entire internal LM or the predictor alone.



\begin{table}[t]
\setlength{\tabcolsep}{7.0 pt}
\centering
\begin{tabular}[c]{c|c|c|c|c|c}
	\hline
	\hline
	\multirow{2}{*}{\begin{tabular}{@{}c@{}} Method
		\end{tabular}} & \multirow{2}{*}{\begin{tabular}{@{}c@{}} Adapted \\ Params 
		\end{tabular}} & \multicolumn{4}{c}{KLD Regularization Weight $\rho$} \\
	\hhline{~~----}	
	& & 0.0 & 0.2 & 0.5 & 0.8  \\
	\hline
    T-T & - & \multicolumn{4}{c}{12.81} \\
	\hline
  \multirow{3}{*}{\begin{tabular}{@{}c@{}} + ILMA 
		\end{tabular}} & ILM & 12.30 & 11.93 & 12.44 & 12.29 \\
	\hhline{~-----}	
     & Predictor & 12.32 & 12.01 & 12.34 & 12.83 \\
	\hhline{~-----}	
     & Joiner & 10.85 & 10.36 & 10.53 & 11.09 \\
	\hline
	\hline
    ILMT & - & \multicolumn{4}{c}{12.55} \\ 			
	\hline
   \multirow{3}{*}{\begin{tabular}{@{}c@{}} + ILMA 
		\end{tabular}} & ILM & 10.34 & 9.67 & 10.17 & 11.59 \\
	\hhline{~-----}	
    & Predictor & 10.47 & 10.02 & 10.42 & 11.50 \\
	\hhline{~-----}	
    & Joiner & 8.96 & \textbf{8.34} & 8.51 & 9.65 \\
	\hline
	\hline
\end{tabular}
\caption{  
The WERs (\%) of T-T models on the \textbf{cross-domain Command \& Control} test set. T-Ts are first trained with a standard transducer loss (T-T) or an ILMT loss, and then adapted with the ILMA loss using Command \& Control \emph{text-only} adaptation data.
Joiner represents $\mathbf{W}_j^\text{NB}, \mathbf{b}^\text{NB}_j$, i.e., the joint network output layer after removing the rows corresponding to the blank.
}
\label{table:command_control}
\vspace{-15 pt}
\end{table}

We then perform ILMA of the ILMT-ed T-T using the text-only data from the Command \& Control adaptation set.
ILMA achieves 24.5\%, 21.8\% and 34.9\% relative WER reductions from the baseline T-T by updating the entire internal LM, the predictor alone or $\{\mathbf{W}_j^\text{NB}, \mathbf{b}_j^\text{NB}\}$ of the joint network, respectively, when $\rho=0.2$. Among the three update schemes, adapting $\{\mathbf{W}_j^\text{NB}, \mathbf{b}_j^\text{NB}\}$ alone performs the best, achieving 13.8\% and 16.8\% relative WER reductions from adapting the whole internal LM and adapting only the prediction network, respectively. It also outperforms ILMT-ed T-T and FTA by 33.5\% and 16.9\% relatively in terms of lower WER, respectively. Moreover, in Table \ref{table:compare_adaptation}, ILMA achieves a 4.2\% relative WER reduction from Shallow Fusion with 46.4\% fewer model parameters and 20.7\% shorter inference time.

\subsubsection{Intra-Domain Evaluation}
\label{sec:exp_ilma_ilmt}

We perform ILMA of the baseline T-T using the Multi-Domain adaptation text.  
However, as in Table \ref{table:voice_search}, adapting the entire internal LM or the predictor alone degrades the WER of the baseline T-T for all regularization weights. Even adapting $\{\mathbf{W}_j^\text{NB}, \mathbf{b}_j^\text{NB}\}$ alone achieves little relative WER reduction from the baseline T-T. 

\begin{table}[h]
\setlength{\tabcolsep}{8.0 pt}
\centering
\begin{tabular}[c]{c|c|c|c|c|c}
	\hline
	\hline
	\multirow{2}{*}{\begin{tabular}{@{}c@{}} Method
		\end{tabular}} & \multirow{2}{*}{\begin{tabular}{@{}c@{}} Adapted \\ Params 
		\end{tabular}} & \multicolumn{4}{c}{KLD Regularization Weight $\rho$} \\
	\hhline{~~----}	
	& & 0.0 & 0.2 & 0.5 & 0.8  \\
	\hline
    T-T & - & \multicolumn{4}{c}{4.23} \\
	\hline
  \multirow{3}{*}{\begin{tabular}{@{}c@{}} + ILMA 
		\end{tabular}} & ILM & 4.31 & 4.31 & 4.32 & 4.29 \\
	\hhline{~-----}	
     & Predictor & 4.28	& 4.30 & 4.30 & 4.31 \\
	\hhline{~-----}	
     & Joiner & 4.19 & 4.19 & 4.19 & 4.18 \\
	\hline
	\hline
    ILMT & - & \multicolumn{4}{c}{4.22} \\ 			
	\hline
   \multirow{3}{*}{\begin{tabular}{@{}c@{}} + ILMA 
		\end{tabular}} & ILM & 4.13 & 4.12 & 4.12 & 4.13 \\
	\hhline{~-----}	
    & Predictor & 4.11 & 4.10 & 4.09 & 4.10 \\
	\hhline{~-----}	
    & Joiner & 4.07 & 4.01 & \textbf{3.95} & 4.00 \\
	\hline
	\hline
\end{tabular}
\caption{  
The WERs (\%) of T-T models on the \textbf{intra-domain Voice Search} test set. T-Ts are first trained with a standard transducer loss (T-T) or an ILMT loss, and then adapted with the ILMA loss using Multi-Domain \emph{text-only} adaptation data.
Joiner represents $\mathbf{W}_j^\text{NB}, \mathbf{b}^\text{NB}_j$, i.e., the joint network output layer after removing the rows corresponding to the blank.
}
\label{table:voice_search}
\vspace{-20 pt}
\end{table}

We then adapt the internal LM of the ILMT-ed T-T using the Multi-Domain text. ILMA achieves 2.6\%, 3.3\% and 6.6\% relative WER reductions from the baseline T-T when updating the entire internal LM, the predictor alone and $\{\mathbf{W}_j^\text{NB}, \mathbf{b}_j^\text{NB}\}$ of the joint network, respectively, when $\rho=0.5$. Adapting $\{\mathbf{W}_j^\text{NB}, \mathbf{b}_j^\text{NB}\}$ performs the best among all three update schemes, with 6.4\% and 4.8\% relative WER reductions from ILMT-ed T-T and FTA, respectively. Furthermore, in Table \ref{table:compare_adaptation}, ILMA achieves a 2.2\% relative WER reduction from Shallow Fusion with 46.4\% fewer model parameters and 25.6\% shorter inference time.

\subsection{Result Analysis}
T-T + ILMA results in Table \ref{table:voice_search} indicate that the ILMA of a T-T trained with standard transducer loss may degrade the ASR performance because minimizing the ILMA loss pushes the internal LM towards a pure NN-LM that is incompatible with the other components of the E2E model. 
By comparing the T-T + ILMA and ILMT + ILMA results in both tables,
we see that the ILMA of a T-T trained with ILMT loss achieves much larger relative WER reductions than that of a T-T trained with a standard transducer loss for both cross-domain and intra-domain evaluations. This shows that the dual-mode internal LM learned via ILMT can be adapted effectively as a standalone LM using \emph{text-only} data while its full functionality as an essential T-T component is well maintained.

ILMT + ILMA results in Tables \ref{table:command_control} and \ref{table:voice_search} suggest that adding a proper KLD regularization ($\rho>0$) always reduces WER by preventing the E2E model from overfitting the adaptation text. 
Furthermore, ILMA by updating only $\{\mathbf{W}_j^\text{NB}, \mathbf{b}_j^\text{NB}\}$ of the joint network shows consistently and significantly better performance than updating the entire internal LM or the prediction network alone. 

Compared to FTA \cite{pylkkonen2021fast}, the proposed ILMA saves one step of output LM component training, achieving \emph{simpler} and \emph{faster} text-only adaptation. Most importantly, as in Table 1, ILMA achieves 4.8\% - 16.9\% relatively lower WERs than FTA. Further, ILMA consistently outperforms Shallow Fusion with significantly reduced model parameters, computational cost and inference time.

\section{Conclusion}
In this work, we propose an internal LM adaptation of the E2E model using text-only data. During ILMA, the internal LM of the E2E model is fine-tuned with adaptation text to minimize a cross-entropy internal LM loss. The internal LM training of the E2E model is performed before ILMA to ensure that the internal LM behaves like a standalone NN-LM while maintaining its functionality as an inseparable component of the E2E model. The KLD regularization is added to the ILMA loss to avoid overfitting. 

Experimented with a 30K-hour trained T-T model, ILMA achieves up to 34.9\% and 6.6\% relative WER reductions from the baseline T-T on cross-domain and intra-domain test sets, respectively. We also show that ILMT is necessary for an effective and robust ILMA. During ILMA, updating only the last linear layer of the joint network consistently achieves the best performance.

\bibliographystyle{IEEEtran}

\bibliography{refs}


\end{document}